\documentclass[11pt]{article}

\usepackage[final]{acl} 
\usepackage{xurl}
\usepackage{times}
\usepackage{latexsym}

\usepackage[T1]{fontenc}

\usepackage[utf8]{inputenc}

\usepackage{inconsolata}

\usepackage{graphicx}

\usepackage{tabularx, booktabs}
\usepackage{enumitem, listings, xcolor}
\usepackage[most]{tcolorbox}

\title{Beyond Reflection: Affirmation as a Promising Behavioral Marker Associated with Quality in Text-Based Counseling}

\author{Michimasa Inaba\\
  The University of Electro-Communications \\ 
  1-5-1, Chofugaoka, Chofu, Tokyo, Japan \\
  \texttt{m-inaba@uec.ac.jp} \\}

\begin{document}
\maketitle
\begin{abstract}
While AI-assisted text-based counseling is gaining attention, it remains empirically unclear which counselor behaviors are associated with higher dialogue quality. Existing research often focuses heavily on Reflection, borrowing frameworks from Motivational Interviewing. To address this gap, we conduct a multi-layered analysis using KokoroChat, a large-scale Japanese text counseling dataset conducted by professional counselors and trainees, newly annotated with counselor strategy tags and client distress levels. 
Our results show that, under the quality indicators used in this study, Affirmation is more consistently associated with session quality than Reflection among the analyzed strategies. Cross-dataset transfer experiments further suggest that this quality signal can be observed to some extent on ESConv, an English dataset with non-expert supporters. These findings provide empirical implications for counselor training and emotional support system design. 
We release the additional KokoroChat annotations and experimental source code at https://github.com/UEC-InabaLab/BeyondReflection.
\end{abstract}

\section{Introduction}
Research on the application of artificial intelligence in mental health support has attracted significant attention \cite{gkotsis2017characterisation,ji2018supervised,nguyen-etal-2022-improving}. AI-assisted support offers several accessibility advantages, such as eliminating geographical barriers, reducing the stigma of seeking help through anonymity, and providing low-cost availability \cite{torous2020digital}. Furthermore, recent advancements in large language models (LLMs) have accelerated research on response generation specifically for text-based counseling \cite{cheng-etal-2022-improving,ChatCounselor,inaba2024,xu-etal-2025-multiagentesc,STAMPsy}.

Despite these advancements, empirical evidence remains limited regarding how a broad range of counselor strategies are comparatively associated with session-level quality indicators in text-based counseling.
Empirically elucidating what counselor behaviors enhance quality in this domain is crucial. Such insights are essential not only for supporting the training of human counselors but also for establishing foundational design principles for AI-assisted support systems.

To address the lack of specific evaluation frameworks, existing studies have frequently referred to the framework of Motivational Interviewing (MI). In the field of MI, behavioral coding systems such as MISC \cite{miller2003manual} and MITI \cite{moyers2016motivational} are well-established, defining behavioral categories represented by OARS (Open questions, Affirmations, Reflections, and Summaries) \cite{miller2012motivational}. Among these, Reflection is considered the core skill, and NLP research has primarily focused on MI behavior classification as well as the generation and detection of Reflections \cite{perez-rosas-etal-2017-understanding,info:doi/10.2196/12529,shen-etal-2020-counseling}. However, MI is a technique specialized for behavior change, such as smoking cessation or reducing alcohol consumption, and its primary objective differs from that of emotional support counseling, which aims to alleviate distress and provide emotional stability.

Despite this difference in objectives, Reflection and Restatement have been explicitly treated as targets for analysis and generation in emotional support dialogue research \cite{liu-etal-2021-towards,cheng-etal-2022-improving}. Furthermore, even in studies targeting online mental health consultations, Reflection and Restatement are analyzed as crucial counselor strategies \cite{oneil-etal-2023-automatic}.
Complementary mixed-methods research on CBT-based online peer support identified active listening, reflective restatements, and opportunities for exploration as factors contributing to support seekers’ perceived empathy, while also reporting discrepancies between human and computational assessments of empathy \cite{10.1145/3613904.3642034}.

Although a previous study has analyzed large-scale text-based counseling data \cite{althoff-etal-2016-large}, focusing primarily on word frequencies and vocabulary, it does not conduct an analysis based on counselor behaviors or strategies.

To address this gap, this study empirically investigates counselor behaviors using KokoroChat \cite{qi-etal-2025-kokorochat}, a large-scale Japanese text-based counseling dataset. KokoroChat includes counselor evaluations provided by the clients, which we utilize as indicators of dialogue quality. Furthermore, we newly annotate KokoroChat with counselor strategy tags and client distress levels to conduct a multi-layered analysis, allowing us to examine associations between counselor behaviors and the client's state.
In addition, to examine whether the obtained findings are specific to KokoroChat, we perform a cross-dataset transfer validation on ESConv \cite{liu-etal-2021-towards}, an English emotional support dialogue dataset, using commonly definable behavioral features. This approach enables us to examine whether the findings are dependent on a single corpus or if they can be observed to some extent across different conditions of culture, language, and expertise.

The contributions of this study are threefold: 
(1) We provide an additional annotation layer for 6,589 KokoroChat sessions, consisting of 11 types of counselor strategy tags and 4-level client distress labels, and publicly release these annotations and the experimental code.
(2) Through a multi-layered analysis, we empirically show that Affirmation, a specific counselor strategy, is more consistently associated with our quality indicators than Reflection among the analyzed strategies.
(3) By performing cross-dataset transfer between two datasets differing in language, cultural background, and counselor expertise, we show that the quality signal associated with Affirmation can be observed to some extent beyond KokoroChat.

\section{Data and Annotation}
\subsection{KokoroChat}

In this study, we conducted our analysis using KokoroChat \cite{qi-etal-2025-kokorochat}, a Japanese text-based counseling dataset. KokoroChat contains one-on-one role-play text counseling sessions conducted by professional counselors or trainees. Each session lasts for 60 minutes, and the dataset includes a total of 6,589 sessions. All sessions are annotated with evaluation scores provided by the speakers who played the role of the client. This evaluation consists of 20 items, each scored from 0 to 5, resulting in a maximum total score of 100.

\subsection{Annotation of Counselor Strategies}
To conduct an analysis focusing on counselor strategies, we newly annotated each counselor utterance with one of the 11 mutually exclusive strategy tags summarized in Table~\ref{tab:strategy_definitions}. Affirmation, Suggest, and Inform were adopted from ESConv \cite{liu-etal-2021-towards} with shortened names. We divided ESConv's Question category into OpenQuestion and ClosedQuestion following the distinction used in Anno-MI \cite{annomi}. Reflection and Paraphrase follow ESConv's corresponding definitions and align with Anno-MI's Complex Reflection and Simple Reflection, respectively. We additionally introduced Backchannel, Greeting, and Thanking because these behaviors occurred frequently in KokoroChat. The rare ESConv Self-disclosure category was included in Other.

The distinction among Affirmation, Reflection, and Paraphrase is central to our analysis. Affirmation positively evaluates the client's strengths, efforts, motivations, abilities, or actions. Reflection verbalizes the client's emotions or underlying personal meaning, whereas Paraphrase restates the factual or propositional content of the client's preceding utterance. Thus, empathizing with or validating an emotion alone is classified as Reflection rather than Affirmation.

\begin{table*}[t]
\centering
\footnotesize
\begin{tabularx}{\textwidth}{@{}p{2.0cm} X p{4.2cm}@{}}
\toprule
\textbf{Tag} & \textbf{Definition} & \textbf{Example utterance} \\
\midrule
OpenQuestion & A question intended to elicit a broad or elaborated response from the client. & ``What has happened recently?'' \\
ClosedQuestion & A yes/no question or one with a restricted answer set, including questions about quantities or time. & ``Did that happen recently?'' \\
Paraphrase & A restatement of the factual or propositional content of the client's preceding utterance, without interpreting emotion. & ``So, what you mean is that ...'' \\
Reflection & A verbalization of the client's emotions or underlying personal meaning intended to clarify feelings or convey empathic understanding. & ``That must have been painful for you.'' \\
Affirmation & A positive evaluation of the client's strengths, efforts, motivations, abilities, or actions; emotional validation alone is insufficient. & ``It is wonderful that you have been able to keep trying that hard.'' \\
Suggest & A proposal of an action, perspective, or solution intended to encourage the client to act or reconsider. & ``How about trying ...?'' \\
Inform & Provision of facts, opinions, resources, or answers without encouraging a specific action. & ``In general, it is said that ...'' \\
Backchannel & A brief signal of attentive listening that contains no substantive summary or reflection. & ``I see.'' \\
Greeting & An utterance whose primary function is greeting or opening the interaction. & ``Hello.'' \\
Thanking & An utterance whose primary function is expressing gratitude. & ``Thank you.'' \\
Other & An utterance not covered above, such as session management, apology, or rare self-disclosure. & ``Please wait a moment.'' \\
\bottomrule
\end{tabularx}
\caption{Counselor strategy annotation scheme for KokoroChat, with definitions and example utterances.}
\label{tab:strategy_definitions}
\end{table*}

The annotation was automatically performed on a total of 306,495 counselor utterances included in KokoroChat using Gemini 2.5 Flash. The actual prompt used for this process is shown in Appendix \ref{sec:appendix_prompt}.

\subsection{Annotation of Distress Levels}

To analyze the extent to which the client's distress was alleviated throughout the session, we annotated the client's utterances on a 4-point scale ranging from 0 (no distress) to 3 (severe distress). The annotation was performed automatically using the Gemini 2.5 Flash model, treating consecutive client utterances as a single block. The actual prompts used are provided in Appendix \ref{sec:appendix_prompt2}.

\subsection{Evaluation of Automatic Annotations}
To validate the automatic annotations, two crowd workers independently annotated a subset of the data. For counselor strategies (318 utterances), Cohen's $\kappa$ between workers was 0.674, and the scores between Gemini and each worker were 0.710 and 0.687. For distress levels (204 blocks), the Quadratic Weighted $\kappa$ between workers was 0.677, and the scores between Gemini and each worker were 0.611 and 0.650. 
All scores indicate substantial agreement.
We further analyzed tag-level error patterns to examine whether LLM annotation errors were systematically concentrated in particular strategy categories; details are provided in Appendix~\ref{app:tag_error_analysis}.

\section{Analysis of Counselor Strategies and Session Quality}
We examined the relationship between counselor behaviors and session quality at the session level to identify behavioral patterns that characterize high-quality dialogues overall. We utilized two complementary quality indicators: categorical and continuous.
For the categorical indicator, we divided the sessions into three tiers (high, mid, and low) based on the top, middle, and bottom thirds of the clients' review scores. For the continuous indicator, we measured the change in client distress, defined as the average distress level in the second half of the session minus that in the first half. A more negative value indicates greater distress reduction. 
Using both indicators allows us to capture quality from multiple perspectives, reflecting both subjective evaluation and state transition.

We also verified that these two indicators were empirically aligned. Client-rated scores were significantly negatively correlated with $\Delta$distress, and mean $\Delta$distress showed a monotonic pattern across the score-based tiers. This suggests that the two indicators capture related but non-identical aspects of dialogue quality; detailed results are provided in Appendix~\ref{app:score_distress}.

\subsection{Tier-based Analysis}

\begin{figure}[t]
    \centering
    \includegraphics[width=\linewidth]{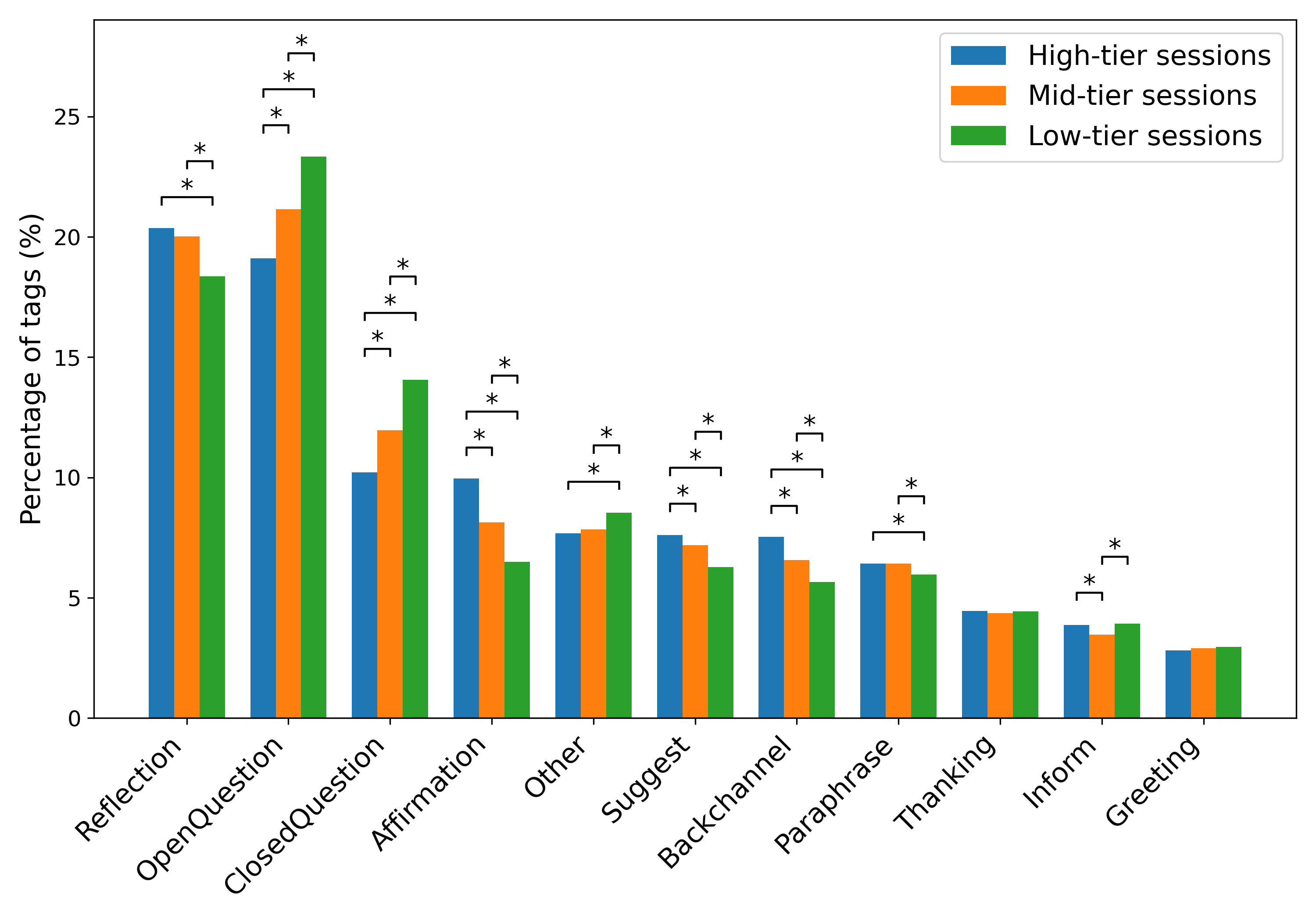}
\caption{Tag usage rates by client-rated score tier. Brackets and asterisks indicate significant pairwise differences between the connected tiers in Bonferroni-corrected chi-square tests at $p < .01$. The significance markers refer to tier differences, not correlations.}
    \label{fig:tag_distribution}
\end{figure}

We calculated the usage rate of each tag across the high, mid, and low client-rated score tiers, and performed chi-square tests followed by Bonferroni-corrected pairwise tier comparisons. Figure~\ref{fig:tag_distribution} shows the results. Exact usage rates, utterance counts, and high--low differences for all tags are provided in Appendix~\ref{app:tier_usage_rates}.
In the figure, brackets and asterisks indicate significant pairwise differences between the connected tiers at the 1\% level after Bonferroni correction. 
Overall, high-tier sessions showed higher use of Affirmation and lower use of questions. Affirmation accounted for 9.9\% of counselor utterances in the high tier and 6.5\% in the low tier, corresponding to an absolute difference of 3.5 percentage points based on the unrounded rates. This pattern suggests that explicit validation and encouragement may be particularly important in text-based counseling, where nonverbal cues of acceptance are unavailable. In contrast, the higher question frequency in lower-tier sessions may reflect a reliance on information gathering rather than more empathic strategies.

\subsection{Distress Level-Based Analysis}

Next, to complement the client-rated tiers with a state-transition-based outcome, we analyzed correlations between distress change and tag usage rates.
We also calculated partial correlations controlling for initial distress, defined as the mean distress level in the first half of the session, to reduce the possibility that the associations merely reflected differences in baseline distress.

\begin{table}[t]
\small
\centering

\begin{tabular}{lrrrr}
\toprule
Tag & $\rho$ & Adj. $p$ & Partial $\rho$ & Adj. $p$ \\
\midrule
Affirmation         & -0.201 & $< .001$ & -0.200 & $< .001$ \\
Suggest             & -0.130 & $< .001$ & -0.119 & $< .001$ \\
Thanking            & -0.040 &   0.025  & -0.050 & $< .001$ \\
Other               & -0.027 &   0.523  & -0.035 &   0.080  \\
Paraphrase          & -0.018 &   1.000  & -0.034 &   0.113  \\
Inform              & -0.014 &   1.000  & -0.001 &   1.000  \\
Backchannel         &  0.003 &   1.000  &  0.019 &   1.000  \\
Reflection          &  0.024 &   0.936  &  0.075 & $< .001$ \\
Greeting            &  0.030 &   0.294  &  0.003 &   1.000  \\
OpenQuestion        &  0.077 & $< .001$ &  0.049 &   0.001  \\
ClosedQuestion      &  0.126 & $< .001$ &  0.117 & $< .001$ \\
\bottomrule
\end{tabular}
\caption{Spearman and partial Spearman correlations between counselor strategy tag usage rates and $\Delta$distress. Partial correlations control for the initial distress level, defined as the mean distress level in the first half of the session. Adjusted $p$-values (Adj. $p$) are Bonferroni-corrected.}
\label{tab:delta_distress_tags_partial}
\end{table}

Table \ref{tab:delta_distress_tags_partial} presents the results. Affirmation was the tag most consistently associated with distress alleviation, showing a significant negative correlation with distress change. This indicates that sessions with higher Affirmation usage are associated with greater distress reduction toward the session's end. Consistent with tier-based analysis, we hypothesize that Affirmation explicitly conveys acceptance and validation, fostering security and ease of speaking, particularly in text-based counseling where nonverbal cues are limited.

Because this analysis is observational, we additionally conducted temporal robustness checks and mixed-effects/Mundlak analyses accounting for counselor-level clustering. These analyses did not eliminate the association, but remain correlational; details are provided in Appendices~\ref{app:temporal_affirmation} and~\ref{app:counselor_random_effects}.

The absolute correlation for Affirmation is modest, and thus Affirmation alone should not be interpreted as strongly explaining session quality. Rather, its importance lies in its comparative consistency across score-based tiers, distress-based analyses, robustness checks, and the cross-dataset transfer discussed later.

\section{Cross-Dataset Transfer}

The analyses thus far have demonstrated that Affirmation is the most consistently associated with quality within KokoroChat. 
We next examined whether the behavioral signal is specific to KokoroChat or if it generalizes to some extent across a different dataset varying in language, culture, and expertise. The objective of this section is to examine whether a quality prediction model trained on KokoroChat (Japanese, trained counselors) yields meaningful predictive signals for ESConv \cite{liu-etal-2021-towards}, an emotional support dialogue dataset (English, non-expert supporters).

\subsection{Experimental Setup}

We trained a logistic regression model on a binary classification task distinguishing the high and low tiers based on KokoroChat's client review scores. We targeted 4,390 sessions with the most distinct quality differences, excluding the mid tier. This exclusion prioritizes evaluating the predictive power of behavioral features between sessions with the clearest quality contrast, as the middle tier exhibits mixed characteristics.
We restricted the features to five dimensions that could be consistently defined across both datasets. Specifically, we used the usage rates of Affirmation, Reflection, Question (combining OpenQuestion and ClosedQuestion), Suggest, and Other. These correspond to ESConv's Affirmation and Reassurance, Reflection of feelings, Question, Providing Suggestions, and Others, respectively.
The mapped categories are semantically aligned with ESConv by design: except for Question, KokoroChat uses the same category definitions as ESConv, with shortened label names. The mapping table is provided in Appendices~\ref{app:tag_mapping}.

After training, we used the model to calculate the probability of being classified as the high tier for each ESConv session. We then computed the Spearman's rank correlation coefficient between this predicted probability and $\Delta$emotion, defined as the difference between ``final\_emotion\_intensity'' and ``initial\_emotion\_intensity'' annotated in ESConv.

\subsection{Results}

The predicted probabilities from the model trained only on KokoroChat showed a significant negative correlation with $\Delta$emotion in ESConv ($\rho=-0.072, p = 0.014$).
Although small, this directionally consistent association indicates that sessions exhibiting behavioral patterns classified as ``high quality'' in KokoroChat tended to show greater emotional improvement in ESConv.

\begin{table}[t]
\small
\centering
\setlength{\tabcolsep}{4pt}
\begin{tabular}{@{}lr@{}}
\toprule
Feature & Coefficient \\
\midrule
Affirmation & $+0.536$ \\
Question    & $-0.421$ \\
Suggest     & $+0.207$ \\
Other       & $-0.149$ \\
Reflection  & $+0.102$ \\
\bottomrule
\end{tabular}
\caption{Regression coefficients in the KokoroChat-to-ESConv transfer model.}
\label{tab:transfer_importance}
\end{table}

Table~\ref{tab:transfer_importance} shows the standardized coefficients of the logistic regression model trained on KokoroChat. Affirmation had the largest positive coefficient ($+0.536$), whereas Question had a large negative coefficient ($-0.421$). Thus, higher Affirmation usage increased the model's log-odds of predicting the high tier, while higher Question usage decreased them, consistent with the tier-based analysis in Section~3.1. Suggest and Reflection also had positive coefficients, although their magnitudes were smaller, whereas Other had a negative coefficient.

Admittedly, the correlation coefficient is small, indicating a modest effect size. However, this transfer setting involves multiple discrepancies: cultural and linguistic differences (Japanese to English), expertise differences (trained counselors to non-expert supporters), and differing evaluation metrics (comprehensive session scores versus emotional improvement scores). The fact that a significant association was observed despite these conditions is crucial in itself, suggesting that the quality signal centered on Affirmation is reproducible across datasets to some extent.

\section{Conclusion}

We analyzed counselor behaviors in text-based counseling using the newly annotated KokoroChat dataset. Affirmation showed a stronger and more consistent association with our quality indicators than Reflection under the present evaluation setting. Cross-dataset transfer to ESConv further suggested that the quality signal associated with Affirmation can be observed to some extent across languages and expertise levels. These findings provide empirical implications for counselor training and emotional support system design.

Future work should move beyond session-level frequencies toward sequence-level modeling of when to affirm, conditioned on the client's preceding utterance, current distress level, and session phase. Such modeling would address the inherent limitation of macro-level behavioral coding and directly inform response generation for emotional support systems. The quality of Affirmation itself also warrants finer-grained analysis: distinguishing affirmations grounded in the client's specific efforts and actions from generic praise is particularly important for LLM-based systems, where indiscriminate positive feedback risks degenerating into sycophancy rather than genuine validation. In addition, our interpretation that explicit verbal Affirmation is especially important in text-based settings, where nonverbal cues of acceptance are unavailable, is itself a testable hypothesis; comparative analyses on spoken or face-to-face counseling corpora would clarify whether the observed signal is modality-specific. Examining how these associations vary across client characteristics and problem domains would further support the development of adaptive support systems that tailor strategy use to individual clients.

\section*{Limitations}

This study has several limitations. First, the tiers in KokoroChat are defined based on client reviews, which are not entirely independent of other metrics derived from the same session. Therefore, the internal analysis of KokoroChat may involve partial circularity between the quality definition and the explanatory variables. However, the transfer validation to ESConv utilizes a different metric on an external dataset, indicating that at least part of our main conclusion is supported independently of this circularity.

Second, although the correlation coefficient in the transfer experiment is significant, its absolute value is small, and we cannot claim strong external validity at this stage. The mapped categories are definitionally aligned with ESConv by design; however, differences in language, dialogue setting, and annotation procedures mean that complete operational equivalence across datasets cannot be assumed. Our study demonstrates that the quality signal centered on Affirmation does not completely collapse across datasets, rather than suggesting that it can be applied to heterogeneous datasets with high accuracy.

Third, although we additionally modeled counselor identity as a random effect and decomposed within- and between-counselor variation in Affirmation use (Appendix~\ref{app:counselor_random_effects}), our analysis still cannot fully isolate all counselor-level characteristics. The additional analysis suggests that the association between Affirmation and session quality is not solely attributable to stable differences among counselors. However, unobserved counselor competencies, such as timing, empathy, and rapport-building skills, may still contribute to the observed association.

Fourth, our analysis relies on automatic annotations generated by an LLM (Gemini 2.5 Flash). Although we confirmed substantial agreement with human annotators and conducted an additional tag-level error analysis (Appendix~\ref{app:tag_error_analysis}), LLM-based annotations inherently contain noise and potential biases.

Fifth, our study design is entirely observational and does not support causal inference. Although our additional temporal analyses (Appendix~\ref{app:temporal_affirmation}) reduce the plausibility of a simple reverse-causality explanation in which counselors use Affirmation mainly after clients have already improved, they do not rule out all possible confounding. Highly skilled counselors might use Affirmation more frequently, but their overall effectiveness could stem from other unmeasured competencies, such as better timing, empathy, or rapport-building skills. Therefore, we interpret Affirmation as a temporally robust behavioral marker associated with session quality, rather than as a causal determinant of client improvement.

Finally, it is crucial to acknowledge that the absolute values of the observed correlations are modest. For example, the correlation between Affirmation usage and $\Delta$distress was $\rho = -0.201$, and the cross-dataset transfer correlation was $\rho = -0.072$.
This indicates that the mere frequency of counselor behavioral strategies explains only a small portion of the variance in session quality. Dialogue quality is likely highly dependent on factors beyond macro-level tag frequencies, such as the precise timing of interventions, contextual appropriateness, the building of rapport, and subtle linguistic nuances. 
While Affirmation emerged as the most consistent indicator among the analyzed tags, our findings highlight the inherent limitations of frequency-based behavioral coding in fully capturing the complexity of counseling quality.

\section*{Ethical Considerations}
This research involves the sensitive domain of mental health and counseling. We strictly adhere to ethical guidelines and outline the following considerations regarding our data, annotation process, and the potential applications of our findings.

First, regarding data privacy and consent, we utilized KokoroChat and ESConv, both of which are publicly available datasets. We used them strictly in accordance with their respective terms of use. KokoroChat consists of role-play sessions conducted by professional counselors and trainees, ensuring that no real patients were involved and no personally identifiable information (PII) of actual clients is included. Similarly, ESConv was collected under appropriate ethical protocols.

Second, regarding the human evaluation of our automatic annotations, we recruited annotators through CrowdWorks (https://crowdworks.jp/), a Japanese crowdsourcing platform. To ensure fair labor practices, we compensated the workers at an hourly rate exceeding the minimum wage in Tokyo, Japan (approximately 7 USD per hour).

Finally, while our findings provide foundational insights for designing AI-assisted emotional support systems, we are fully aware of the significant ethical risks associated with deploying fully autonomous AI in mental health care. Mental health interventions are highly sensitive; inappropriate or harmful AI-generated responses could exacerbate a client's distress. Therefore, the behavioral markers identified in this study (e.g., Affirmation) are primarily intended to support the training of human counselors and to inform human-in-the-loop AI systems, ensuring that safety and ethical standards are prioritized.

\bibliography{custom}

\clearpage

\appendix

\section{Design Principles for Counselor Strategy Tags}
\label{sec:appendix_tag}
The tags used in this study are based on the eight tags proposed in ESConv. 
Among them, ``Affirmation and Reassurance'', ``Providing Suggestions'', and ``Information'' from ESConv were adopted directly with shortened names: Affirmation, Suggest, and Inform, respectively.
Regarding ``Question'', we adopted the sub-types defined in Anno-MI, splitting it into two distinct tags: OpenQuestion and ClosedQuestion.
In ESConv, ``Reflection of Feelings'' and ``Restatement or Paraphrasing'' correspond to the sub-types ``Complex Reflection'' and ``Simple Reflection'' in Anno-MI, respectively. We mapped these corresponding concepts and adopted them as Reflection and Paraphrase in our study.
Furthermore, based on our analysis of KokoroChat, we newly added three tags: Backchannel, Greeting, and Thanking. These behaviors appeared frequently in KokoroChat, although they fall under the ``Other'' or ``Others'' category in ESConv and Anno-MI.
Initially, we considered ``Self-disclosure'' from ESConv as an independent tag; however, since its frequency in KokoroChat was found to be extremely low, we decided to merge it into the Other tag.

\section{Prompts Used for Strategy Tag Annotation}
\label{sec:appendix_prompt}
\begin{figure*}[t]
    \centering
    \includegraphics[width=\linewidth]{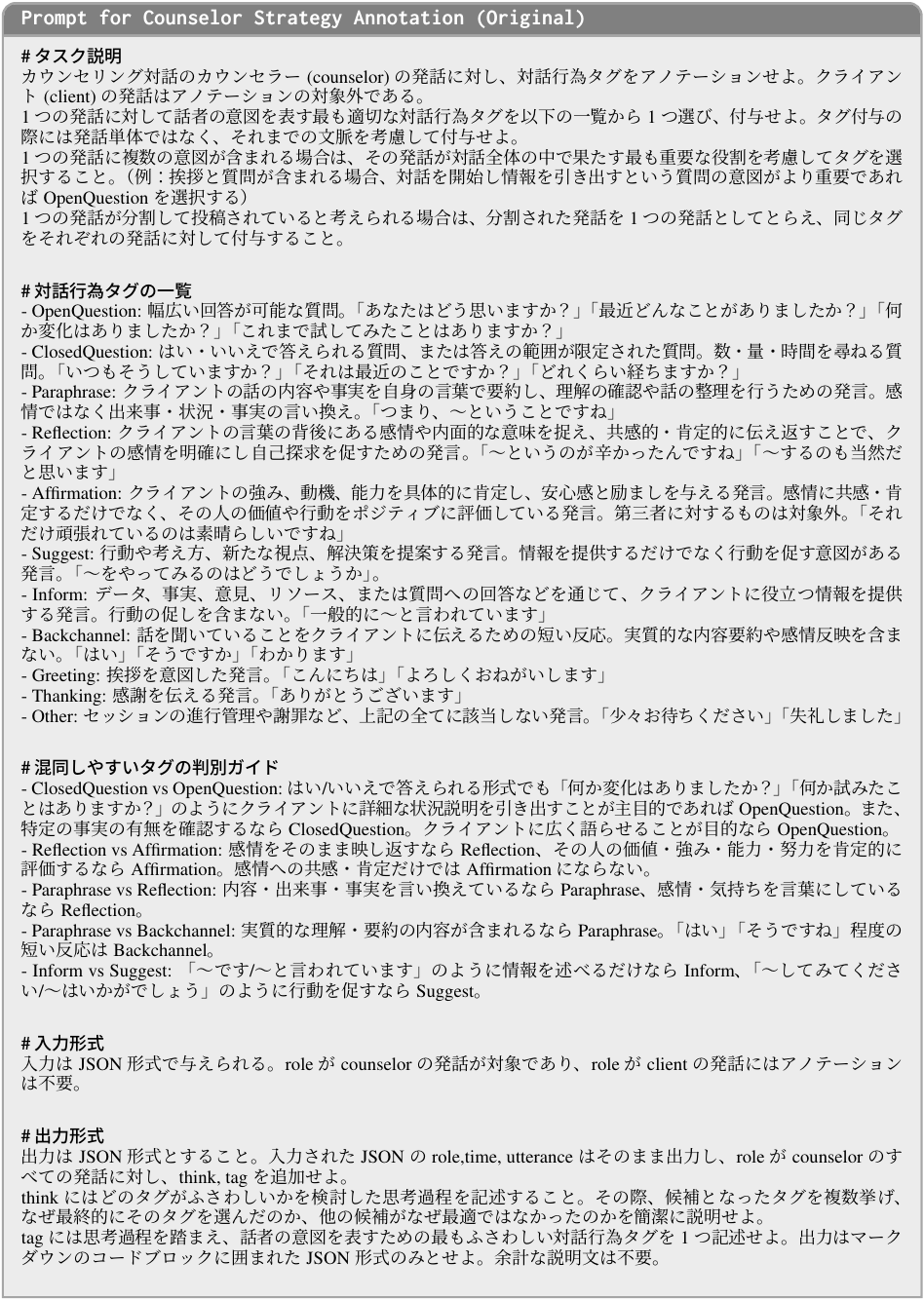}
\caption{Prompt for counselor strategy tag annotation}
    \label{fig:tag_prompt_ja}
\end{figure*}

\begin{figure*}[t]
    \centering
    \includegraphics[width=\linewidth]{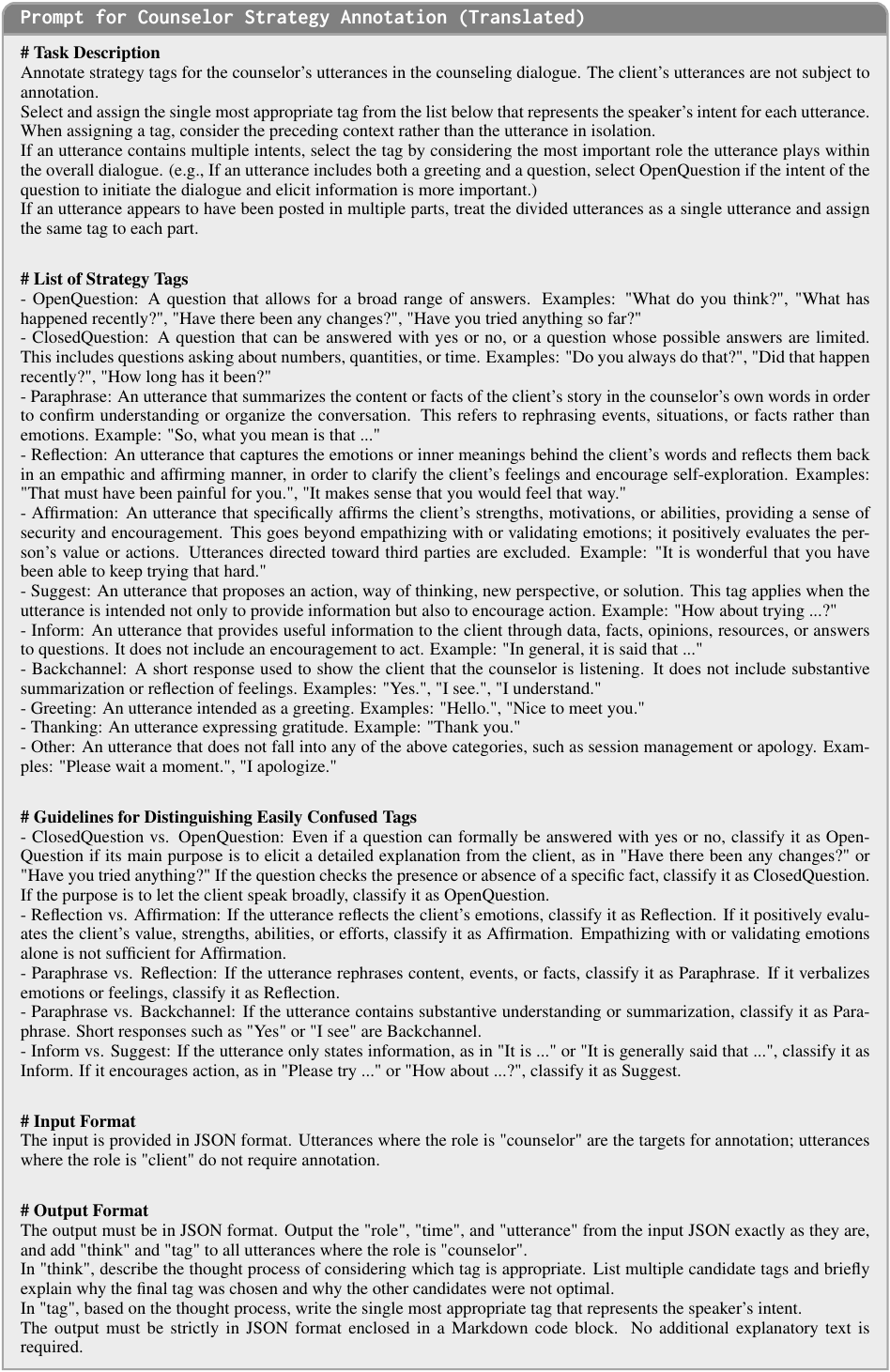}
\caption{English translation of the prompt for counselor strategy tag annotation}
    \label{fig:tag_prompt_en}
\end{figure*}

Figures~\ref{fig:tag_prompt_ja} and~\ref{fig:tag_prompt_en} show the prompt used for counselor strategy tag annotation. The original Japanese prompt in Figure~\ref{fig:tag_prompt_ja} was used for the actual LLM annotation, because all KokoroChat dialogues are in Japanese. Figure~\ref{fig:tag_prompt_en} provides an English translation to make the tag definitions, disambiguation rules, and output format accessible to readers.

\section{Prompts Used for Distress Level Annotation}
\label{sec:appendix_prompt2}
\begin{figure*}[t]
    \centering
    \includegraphics[width=\linewidth]{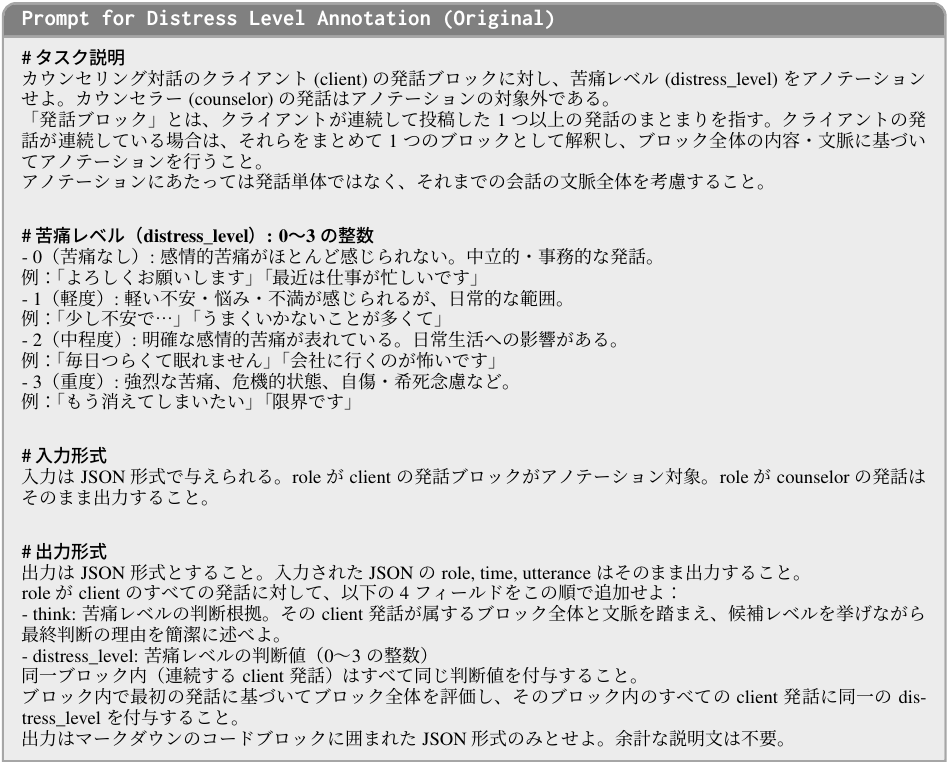}
\caption{Prompt for distress level annotation}
    \label{fig:dis_prompt_ja}
\end{figure*}

\begin{figure*}[t]
    \centering
    \includegraphics[width=\linewidth]{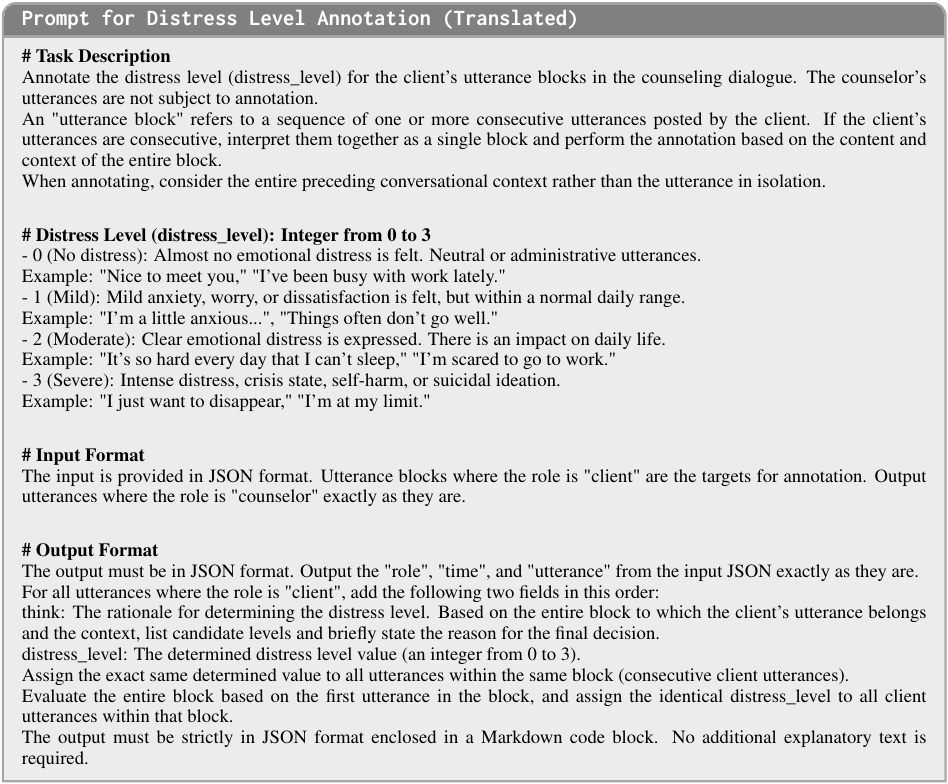}
\caption{English translation of the prompt for distress level annotation}
    \label{fig:dis_prompt_en}
\end{figure*}

Figures~\ref{fig:dis_prompt_ja} and~\ref{fig:dis_prompt_en} show the prompt used for distress-level annotation. As with the strategy tag annotation, the original Japanese prompt in Figure~\ref{fig:dis_prompt_ja} was used for the actual LLM annotation. The English translation in Figure~\ref{fig:dis_prompt_en} is provided to clarify the four-level distress scale, the block-level annotation procedure, and the required JSON output format.

\section{Tag-Level Error Analysis of Automatic Strategy Annotations}
\label{app:tag_error_analysis}

Because the counselor strategy annotations used in this study were automatically produced by Gemini 2.5 Flash, we conducted an additional tag-level error analysis to examine whether annotation errors were randomly distributed or systematically concentrated in particular strategy categories. We compared the LLM annotations with two independent human annotators on the same validation subset used in Section~2.4, consisting of 318 counselor utterances from five sessions. The two human annotators are referred to as Annotator 1 and Annotator 2.

Overall agreement was substantial. Cohen's $\kappa$ was 0.710 between the LLM and Annotator 1, 0.687 between the LLM and Annotator 2, and 0.674 between the two human annotators. Thus, the agreement between the LLM and each human annotator was comparable to, or slightly higher than, the agreement between human annotators.

To examine systematic error patterns, we calculated tag-level precision, recall, and F1 scores by treating the LLM annotation as the prediction and each human annotation as the reference. The scores in Table~\ref{tab:tag_error_analysis} are averaged across the two human references. The results indicate that annotation errors were not uniformly distributed across tags. The lowest F1 score was observed for Other, suggesting that residual or miscellaneous utterances were the most difficult to classify consistently. Moderate performance was observed for ClosedQuestion, Suggest, Paraphrase, Greeting, Backchannel, OpenQuestion, and Thanking. In contrast, Inform, Reflection, and Affirmation showed relatively high F1 scores.

\begin{table}[t]
\centering
\small
\begin{tabular}{lrrrr}
\hline
Tag & Support & Precision & Recall & F1 \\
\hline
Other & 29 & 0.560 & 0.486 & 0.520 \\
ClosedQuestion & 46 & 0.714 & 0.650 & 0.666 \\
Suggest & 14 & 0.579 & 0.790 & 0.667 \\
Paraphrase & 36 & 0.750 & 0.675 & 0.710 \\
Greeting & 8 & 0.650 & 0.817 & 0.721 \\
Backchannel & 14 & 0.658 & 0.870 & 0.744 \\
OpenQuestion & 64 & 0.816 & 0.758 & 0.771 \\
Thanking & 14 & 0.786 & 0.789 & 0.771 \\
Affirmation & 10 & 0.679 & 1.000 & 0.804 \\
Inform & 27 & 0.827 & 0.796 & 0.811 \\
Reflection & 56 & 0.792 & 0.862 & 0.823 \\
\hline
\end{tabular}
\caption{Tag-level precision, recall, and F1 scores of LLM strategy annotations, averaged over Annotator 1 and Annotator 2 as references. Support denotes the average number of utterances assigned to each tag by the two human annotators.}
\label{tab:tag_error_analysis}
\end{table}

The most frequent confusions were concentrated in semantically adjacent categories. For example, OpenQuestion and ClosedQuestion were sometimes confused, and similar ambiguity was also observed between the two human annotators. This suggests that part of the error reflects the intrinsic ambiguity of distinguishing question subtypes rather than an LLM-specific failure. Another frequent confusion involved Paraphrase and Reflection, which are also conceptually close because both involve responding to the client's preceding content.

The error pattern for Affirmation is particularly important for the main findings of this study. The LLM achieved perfect recall for Affirmation with respect to both human annotators: all utterances labeled as Affirmation by Annotator 1 or Annotator 2 were also labeled as Affirmation by the LLM. However, the precision for Affirmation was lower than its recall. The false positives mainly consisted of utterances labeled by humans as Reflection, Paraphrase, or Thanking. Thus, the LLM did not show evidence of systematically under-detecting Affirmation, but it did show a tendency to classify some neighboring supportive responses as Affirmation.

This pattern should be considered when interpreting the downstream analyses. The main association involving Affirmation is unlikely to be explained by a failure to detect true Affirmation utterances, because recall was high. At the same time, the moderate precision indicates that the estimated Affirmation usage rate may include a small number of neighboring supportive strategies. Therefore, our conclusions should be interpreted as applying to automatically identified Affirmation-like behavior, rather than to a perfectly separated manual category. This limitation does not eliminate the main finding, but it reinforces our cautious interpretation of Affirmation as a behavioral marker associated with session quality rather than as an isolated causal mechanism.

\section{Relationship Between Client-Rated Scores and Distress Change}
\label{app:score_distress}

To clarify the relationship between the two quality indicators used in Section~3, we conducted an additional analysis between client-rated review scores and $\Delta$distress. The client-rated score was significantly negatively correlated with $\Delta$distress ($\rho = -0.269$, $p < .001$, $n = 6{,}589$). Since lower $\Delta$distress indicates greater distress reduction, this result suggests that higher-rated sessions tended to involve greater reductions in client distress.

Mean $\Delta$distress also showed a monotonic pattern across the score-based tiers. High-tier sessions showed the largest reduction ($M = -0.204$, $SD = 0.552$, $n = 2{,}168$), mid-tier sessions showed a smaller reduction ($M = -0.053$, $SD = 0.545$, $n = 2{,}199$), and low-tier sessions showed an increase in distress ($M = 0.142$, $SD = 0.572$, $n = 2{,}222$).

These results indicate that the two indicators are aligned in the expected direction. At the same time, the modest correlation suggests that they are not interchangeable. Client-rated scores likely reflect broader subjective aspects of session quality, such as rapport, perceived support, and overall satisfaction, whereas $\Delta$distress captures short-term emotional change within the session. We therefore treat the two indicators as complementary measures of dialogue quality.

\section{Exact Tag Usage Rates by Client-Rated Score Tier}
\label{app:tier_usage_rates}

To supplement Figure~\ref{fig:tag_distribution}, Table~\ref{tab:tier_usage_rates} reports the exact usage rates and utterance counts of each counselor strategy tag by client-rated score tier. Usage rates were calculated over all counselor utterances within each tier. The final column reports the difference between the high- and low-tier usage rates, providing an estimate of the practical magnitude of the tier differences.

\begin{table*}[t]
\centering
\small
\begin{tabular}{lcccc}
\toprule
Tag & High (\%) & Mid (\%) & Low (\%) & $\Delta$ (H$-$L) \\
\midrule
Reflection$^*$      & 20.4 (n=23,257) & 20.0 (n=20,207) & 18.4 (n=16,775) & +2.0 \\
OpenQuestion$^*$    & 19.1 (n=21,825) & 21.1 (n=21,342) & 23.3 (n=21,319) & -4.2 \\
ClosedQuestion$^*$  & 10.2 (n=11,663) & 12.0 (n=12,064) & 14.1 (n=12,833) & -3.8 \\
Affirmation$^*$     &  9.9 (n=11,366) &  8.1 (n=8,209)  &  6.5 (n=5,929)  & +3.5 \\
Other$^*$           &  7.7 (n=8,774)  &  7.8 (n=7,914)  &  8.5 (n=7,797)  & -0.9 \\
Suggest$^*$         &  7.6 (n=8,695)  &  7.2 (n=7,256)  &  6.3 (n=5,736)  & +1.3 \\
Backchannel$^*$     &  7.5 (n=8,607)  &  6.6 (n=6,621)  &  5.7 (n=5,170)  & +1.9 \\
Paraphrase$^*$      &  6.4 (n=7,343)  &  6.4 (n=6,485)  &  6.0 (n=5,440)  & +0.5 \\
Thanking            &  4.5 (n=5,084)  &  4.4 (n=4,407)  &  4.4 (n=4,052)  & +0.0 \\
Inform              &  3.9 (n=4,428)  &  3.5 (n=3,490)  &  3.9 (n=3,576)  &  0.0 \\
Greeting            &  2.8 (n=3,202)  &  2.9 (n=2,931)  &  3.0 (n=2,698)  & -0.2 \\
\bottomrule
\end{tabular}
\caption{Exact counselor strategy tag usage rates by client-rated score tier. Percentages are calculated over all counselor utterances within each tier; counts in parentheses indicate the number of utterances assigned to each tag. The total numbers of counselor utterances were 114,244 for the high tier, 100,926 for the mid tier, and 91,325 for the low tier. $\Delta$ (H$-$L) denotes the high-tier usage rate minus the low-tier usage rate. $^*$ indicates a significant high-vs.-low difference after Bonferroni correction in chi-square tests ($p < .01$).}
\label{tab:tier_usage_rates}
\end{table*}

\section{Temporal Robustness Checks for Affirmation}
\label{app:temporal_affirmation}

Because our study is observational, the associations reported in Section~3 cannot establish a causal effect of Affirmation. In particular, a possible reverse-causality explanation is that counselors may have used more Affirmation only after clients had already improved. To examine this possibility, we conducted additional temporal analyses.

Table~\ref{tab:temporal_affirmation} summarizes the main temporal robustness checks. First, we examined whether Affirmation used in the early part of the session was already associated with session outcomes. Early-session Affirmation was significantly associated with both distress reduction and overall client-rated scores. Specifically, early-session Affirmation was negatively correlated with $\Delta$distress ($\rho = -0.101$, $p < .001$) and positively correlated with client-rated scores ($\rho = 0.126$, $p < .001$). Since lower $\Delta$distress indicates greater distress reduction, these results suggest that the association between Affirmation and session quality was already present before the late phase of the session. Although late-session Affirmation showed stronger associations, the early-session results indicate that the finding is not simply a byproduct of counselors increasing Affirmation only after client improvement.

Second, we examined the reverse-direction relationship between early distress and later Affirmation usage. Early distress did not significantly predict later Affirmation usage in a simple correlation analysis ($\rho = 0.008$, $p = .499$). We do not interpret this result as evidence that early distress has no effect on later counselor behavior. However, it provides no clear support for the explanation that lower early distress leads to greater later use of Affirmation.

We further conducted turn-level analyses predicting whether the next counselor turn was Affirmation. In a mixed-effects logistic regression model with a counselor-level random intercept, current client distress positively predicted subsequent Affirmation use after controlling for session phase and session length (OR = 1.068, 95\% CI [1.050, 1.086], $p < .001$). A generalized estimating equation analysis clustered by session yielded the same conclusion (OR = 1.105, $p < .001$). Thus, higher, rather than lower, current distress was followed by a greater likelihood of Affirmation, which is inconsistent with a simple reverse-causality explanation in which counselors use Affirmation mainly after clients have already improved.

\begin{table*}[t]
\centering
\small
\begin{tabular}{llcc}
\hline
Predictor & Outcome / Model & Effect & $p$ \\
\hline
Early-session Affirmation & $\Delta$distress & $\rho = -0.101$ & $< .001$ \\
Early-session Affirmation & Client-rated score & $\rho = 0.126$ & $< .001$ \\
Early distress & Later Affirmation usage & $\rho = 0.008$ & $.499$ \\
Current distress & Next-turn Affirmation (mixed logit) & OR = 1.068 & $< .001$ \\
Current distress & Next-turn Affirmation (GEE) & OR = 1.105 & $< .001$ \\
\hline
\end{tabular}
\caption{Temporal robustness checks for the relationship between Affirmation, client distress, and session quality. Lower $\Delta$distress indicates greater distress reduction. The 95\% CI for the mixed-effects logistic model was [1.050, 1.086]. GEE was clustered by session.}
\label{tab:temporal_affirmation}
\end{table*}

Third, we decomposed Affirmation usage by session phase. In the three-phase mixed-effects model, middle- and late-phase Affirmation were significantly associated with $\Delta$distress, with the strongest association in the late phase. For client-rated scores, Affirmation was significantly associated with quality across early, middle, and late phases. These results suggest that the association between Affirmation and session quality is temporally heterogeneous rather than merely a session-level frequency artifact.

These analyses do not establish a causal effect of Affirmation and do not rule out all possible confounding. Nevertheless, they reduce the plausibility of the simple explanation that Affirmation is used mainly after clients have already improved. We therefore interpret Affirmation as a temporally robust behavioral marker associated with session quality, rather than as a causal determinant of client improvement.

\section{Counselor-Level Random Effects and Mundlak Decomposition}
\label{app:counselor_random_effects}

\begin{table*}[t]
\centering
\small
\begin{tabular}{llcccc}
\hline
Analysis & Outcome & Estimate & SE & 95\% CI & $p$ \\
\hline
Null mixed-effects model & $\Delta$distress & ICC = 0.028 & -- & -- & -- \\
Null mixed-effects model & Client-rated score & ICC = 0.110 & -- & -- & -- \\
Random-intercept model & $\Delta$distress & $-1.213$ & -- & -- & $< .001$ \\
Random-intercept model & Client-rated score & $85.517$ & -- & -- & $< .001$ \\
Mundlak within-counselor component & $\Delta$distress & $-1.277$ & -- & -- & $< .001$ \\
Mundlak within-counselor component & Client-rated score & $88.035$ & -- & -- & $< .001$ \\
Mundlak between-counselor component & $\Delta$distress & $-0.773$ & $0.300$ & [$-1.360$, $-0.186$] & $.010$ \\
Mundlak between-counselor component & Client-rated score & $+51.828$ & $13.891$ & [$24.601$, $79.055$] & $< .001$ \\
\hline
\end{tabular}
\caption{Additional analyses accounting for counselor-level clustering.
Coefficients for Affirmation usage are based on rates ranging from 0 to 1
and therefore represent the expected outcome change associated with a
full-unit increase in the Affirmation usage rate. Lower $\Delta$distress
indicates greater distress reduction.}
\label{tab:counselor_random_effects}
\end{table*}

\begin{table*}[t]
\centering
\small
\begin{tabular}{lll}
\hline
KokoroChat tag & ESConv tag & Note \\
\hline
Affirmation & Affirmation and Reassurance & Same definition; shortened label \\
Reflection & Reflection of feelings & Same definition; shortened label \\
OpenQuestion + ClosedQuestion & Question & Merged for transfer \\
Suggest & Providing Suggestions & Same definition; shortened label \\
Other & Others & Residual category \\
\hline
\end{tabular}
\caption{Mapping between KokoroChat strategy tags and ESConv strategy tags used in the cross-dataset transfer experiment.}
\label{tab:tag_mapping}
\end{table*}

A possible concern is that the observed association between Affirmation and session quality may reflect stable differences among counselors rather than session-level effects of Affirmation. For example, counselors who generally use more Affirmation may also possess other unobserved competencies, such as stronger rapport-building skills or better timing. To address this concern, we conducted additional mixed-effects analyses with counselor identity modeled as a random intercept.

We first estimated null models to quantify the extent of counselor-level clustering. The intraclass correlation coefficients (ICCs) were 0.028 for $\Delta$distress and 0.110 for client-rated score. These values indicate that counselor-level heterogeneity was small for distress change but non-negligible for subjective client ratings.

We then fitted mixed-effects models predicting each outcome from Affirmation usage while including counselor-level random intercepts. For $\Delta$distress, the model additionally controlled for initial distress, defined as the mean distress level in the first half of the session. The association between Affirmation and $\Delta$distress remained significant after accounting for counselor-level clustering (coefficient = $-1.213$, $p < .001$). The association between Affirmation and client-rated score also remained significant (coefficient = $85.517$, $p < .001$). These results indicate that the main finding was not eliminated by modeling stable differences among counselors.

To further distinguish within-counselor and between-counselor variation, we conducted a Mundlak decomposition of Affirmation usage.
Specifically, Affirmation usage was decomposed into a within-counselor component, representing deviations from each counselor's own mean Affirmation usage, and a between-counselor component, representing each counselor's mean Affirmation usage across sessions.
The within-counselor component remained significant for both $\Delta$distress (coefficient = $-1.277$, $p < .001$) and client-rated score (coefficient = $88.035$, $p < .001$). This indicates that, even for the same counselor, sessions with higher Affirmation usage tended to show better outcomes.

The between-counselor coefficient was $-0.773$ for $\Delta$distress (SE = $0.300$, 95\% CI [$-1.360$, $-0.186$], $p = .010$) and $+51.828$ for client-rated score (SE = $13.891$, 95\% CI [$24.601$, $79.055$], $p < .001$). Thus, counselors with higher mean Affirmation rates tended to show greater distress reduction and higher average client ratings. Because usage rates range from 0 to 1, these coefficients correspond to a full-unit increase in a counselor's mean Affirmation rate.

Table~\ref{tab:counselor_random_effects} summarizes the results of these additional analyses. Overall, the association involving Affirmation is unlikely to be solely attributable to fixed differences among counselors. However, the between-counselor associations remain observational and may also reflect other stable counselor-level competencies. More generally, these analyses remain correlational and do not establish a causal effect of Affirmation.

\section{Mapping Between KokoroChat and ESConv Tags}
\label{app:tag_mapping}

Table~\ref{tab:tag_mapping} summarizes the mapping used in the cross-dataset transfer experiment. The mapped KokoroChat tags were designed to follow the corresponding ESConv definitions, except for Question. ESConv defines Question as a single category, whereas KokoroChat separates it into OpenQuestion and ClosedQuestion following Anno-MI. Therefore, these two KokoroChat tags were merged when transferring the model to ESConv. Other labels differ mainly in naming convention: for example, KokoroChat uses Affirmation as a shortened label for ESConv's Affirmation and Reassurance, and Suggest for ESConv's Providing Suggestions. The full definitions used for KokoroChat annotation are provided in the annotation prompt in Appendix~\ref{sec:appendix_prompt}.

\end{document}